\documentclass[conference]{IEEEtran}
\IEEEoverridecommandlockouts

\usepackage{cite}
\usepackage{amsmath,amssymb,amsfonts}
\usepackage{algorithmic}
\usepackage{graphicx}
\usepackage{booktabs,multirow,siunitx}
\usepackage[table]{xcolor}
\usepackage{textcomp}
\usepackage{xcolor}
\usepackage{booktabs}
\usepackage{array}

\newcolumntype{L}[1]{>{\raggedright\arraybackslash}p{#1}}
\usepackage{marvosym}
\def\BibTeX{{\rm B\kern-.05em{\sc i\kern-.025em b}\kern-.08em
    T\kern-.1667em\lower.7ex\hbox{E}\kern-.125emX}}
\usepackage{xurl}
\begin{document}

\bstctlcite{IEEEexample:BSTcontrol}

\bstctlcite{IEEEBSTcontrol}
\title{PS-DLO: Property- and Spatially Aware Deformable Linear Object Manipulation Policy}




\author{
Mengdi Li$^{1,*}$,
Boxian Yao$^{2,*,\dagger}$,
Xiyu Zhu$^{3,*}$,
Yunfan Lou$^{2}$,
Yifan Yang$^{2}$,\\[2pt]
Yuanpeng He$^{5}$,
Xuan Han$^{2}$,
Tao Liu$^{4}$,
and Zhibo Pang$^{2}$\textsuperscript{\Letter}

\thanks{$^{*}$Equal contribution, $^{\dagger}$Project leader, \textsuperscript{\Letter}Corresponding author.}
\thanks{$^{1}$Mengdi Li is with the School of Computer Science and Technology,
University of Chinese Academy of Sciences, Beijing, China.
{\tt\small limengdi24@mails.ucas.ac.cn}}
\thanks{$^{2}$Boxian Yao, Yunfan Lou, Yifan Yang, Xuan Han, and Zhibo Pang
are with the School of Advanced Manufacturing and Robotics,
Peking University, Beijing, China.
{\tt\small yaoboxian26@stu.pku.edu.cn,
e1373933@u.nus.edu,
2601213890@stu.pku.edu.cn,
xuanhan@berkeley.edu,
zhibo.pang@pku.edu.cn}}
\thanks{$^{3}$Xiyu Zhu is with the College of Engineering,
Peking University, Beijing, China.
{\tt\small xiyu070108@gmail.com}}
\thanks{$^{4}$Tao Liu is with Space Engineering University,
Beijing, China.
{\tt\small 18300602501@163.com}}
\thanks{$^{5}$Yuanpeng He is with the School of Computer Science,
Peking University, Beijing, China.
{\tt\small heyuanpeng@stu.pku.edu.cn}}

\thanks{This work is funded by the Beijing Natural Science Foundation International Scientists Project IS26042 and the State Key Laboratory of High-end Heavy-load Robots (Open Fund Project No. HHR2026010110).}

}

\maketitle

\begin{abstract}
Deformable linear objects (DLO) pose a coupled perception and control challenge: their slender geometry makes precise localization difficult, while their deformations depend on object-specific physical properties. To address this challenge, we propose PS-DLO, a property- and spatially aware DLO manipulation policy. First, contrastive image–text learning maps the initial observation to categorical descriptions of DLO properties, providing explicit physical and structural priors for action generation. Second, SAM3 extracts mask-guided features from future observations, which are distilled into the policy’s action expert to encode task-relevant DLO regions and their subsequent configurations. We evaluate PS-DLO on nine real-world tasks across single-arm and bimanual platforms, training a shared multi-task policy for each embodiment. PS-DLO achieves an average success rate of 68.9\%, exceeding the strongest baseline by 20.0 percentage points. In generalization experiments, PS-DLO achieves higher success rates than the baseline in six of seven visual-shift conditions on butterfly-knot untying. These results demonstrate that combining object-property priors with future spatial supervision improves multi-task DLO manipulation and robustness to visual variation.

\end{abstract}



\section{Introduction}

Recent advances in robotic manipulation of DLO, such as cables, ropes, and wires, have enabled
increasingly complex tasks, including routing, threading, and
knot manipulation~\cite{caporali2026survey}. Despite their simple
appearance, DLO exhibit distinctive geometric and physical
characteristics that complicate manipulation. Their slender
structures, frequent crossings, and self-occlusions make
interaction-relevant segments difficult to distinguish from
surrounding scene content. Meanwhile, their configurations evolve
under contact, with deformation governed by object-dependent
properties such as bending stiffness and extensibility.
Variations in these properties can require different grasping
and motion strategies even for the same task goal. Reliable
DLO manipulation therefore requires policies that accurately
localize relevant object regions while adapting their actions
to the physical properties of the manipulated object.

Previous DLO research has demonstrated the value of geometric modeling and property-conditioned control within specialized manipulation pipelines~\cite{yu2023coarse, li2024flexibility}.
Vision-Language-Action models (VLA) offer a promising route toward shared multi-task policies by combining pretrained multi-modal representations with robot action learning~\cite{kim2024openvla, black2025pi05}. However, adapting these models through action supervision alone leaves the use of spatial and physical cues largely implicit. Demonstrated actions specify the desired behavior but do not directly supervise which object regions and property cues should be preserved in the representations used to generate that behavior.
Although such information may be available in visual observations, its consistent use across DLO configurations and instances is not explicitly encouraged. This raises a central question: How can we explicitly guide a shared VLA policy to attend to interaction-relevant regions while incorporating object-dependent physical information into action generation?

\begin{figure}[t]
    \centering
    \includegraphics[width=\columnwidth]{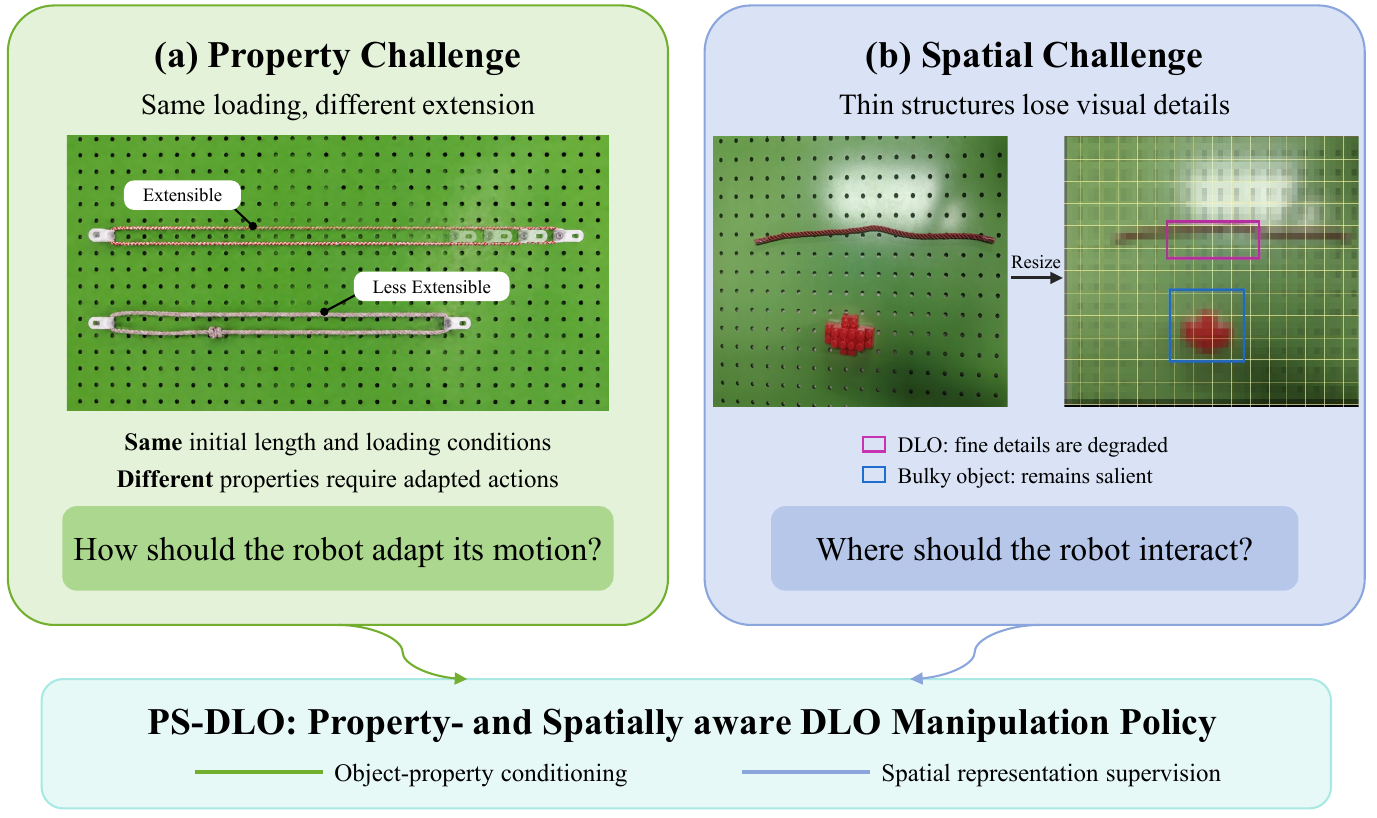} 
    \caption{
    \textbf{Property and spatial challenges in DLO manipulation.}
    (a) DLO with the same initial length exhibit different extensions under identical loading conditions, illustrating the need to adapt actions to object-specific physical properties. (b) Image downsampling degrades fine details of slender DLO while bulky objects remain visually salient, making interaction-relevant regions difficult to localize. These complementary challenges motivate PS-DLO, which combines object-property conditioning with future spatial representation supervision to guide action learning.}
    \label{fig:teasor} 
\end{figure}

To address this challenge (Fig.~\ref{fig:teasor}), we propose PS-DLO, a property- and spatially aware manipulation policy for DLO that integrates object-property conditioning and future spatial representation distillation into action learning. First, contrastive image--text learning aligns initial observations with categorical descriptions of DLO properties, providing explicit physical and structural information. The property prompt is combined with the task prompt to condition action generation throughout the episode. Second, a Teacher Encoder is pretrained with action supervision to adapt mask-guided features extracted by frozen SAM3~\cite{carion2025sam3} from future observations into action-relevant spatial representations. These representations are then distilled into learnable DLO queries within the student policy's action expert. To continually refine the supervisory representations, we introduce a teacher assistant (TA), initialized from the pretrained teacher and optimized through a separate auxiliary action head. The teacher is updated through an exponential moving average (EMA) of the TA encoder's parameters, incorporating ongoing action-driven refinement while providing temporally smoothed targets for student alignment. Together, object-property conditioning and this teacher--assistant mechanism guide policy learning with explicit object-level priors and spatial supervision informed by demonstrated actions. Future observations, SAM3, both teacher-side encoders, and the auxiliary action head are required only during training.

\begin{figure*}[t]
    \centering
    \includegraphics[width=0.98\textwidth]
    {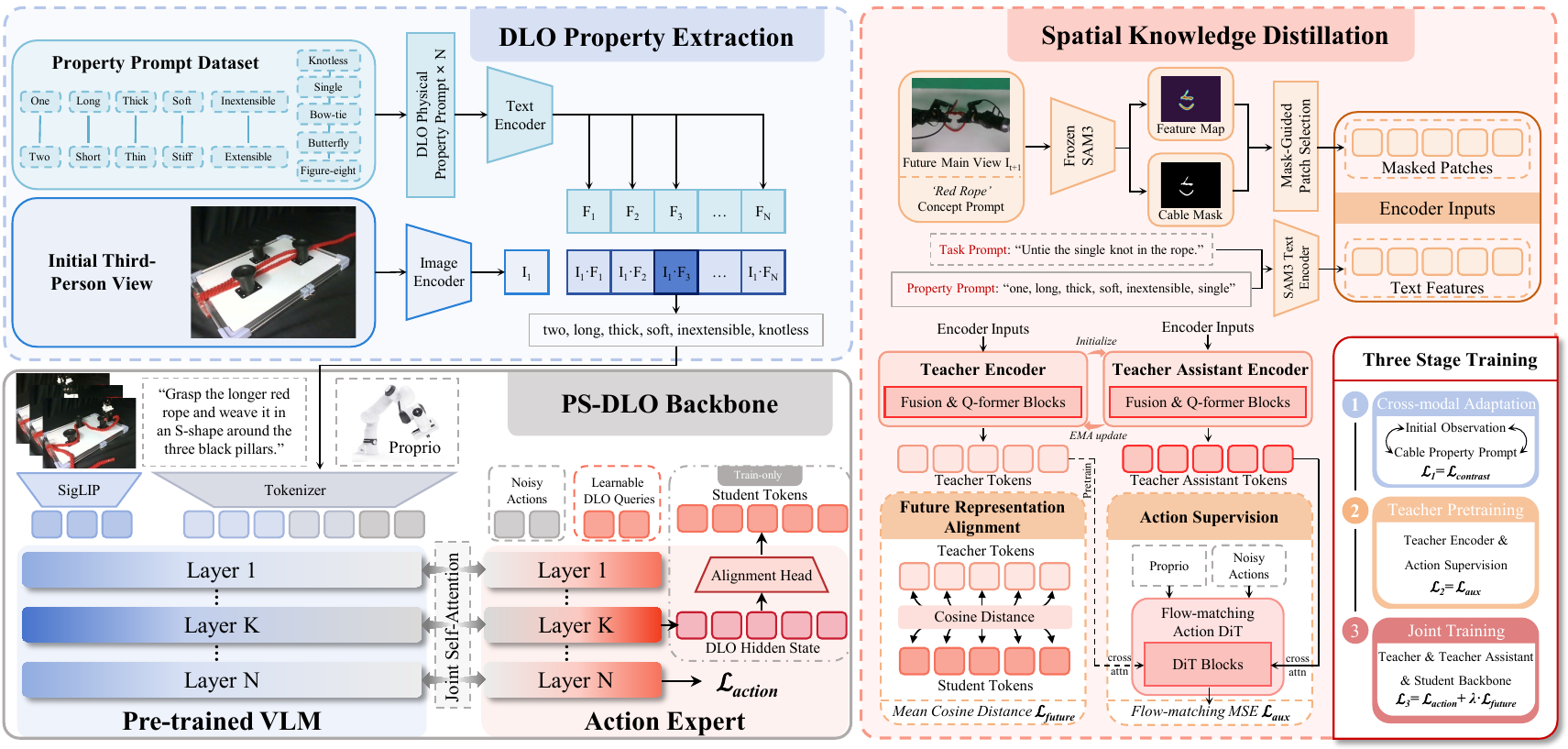}
    \caption{\textbf{Overview of PS-DLO}. The framework consists of DLO property extraction and future spatial knowledge distillation. The property extractor maps the initial observation to a categorical property description for policy conditioning, while the distillation branch uses SAM3-guided features from future observations to provide spatial supervision for the action expert through a teacher--assistant scheme. Training proceeds in three stages: contrastive property-extractor adaptation, teacher pretraining with auxiliary action supervision, and joint training with teacher--student alignment.}
    \label{fig:architecture}
\end{figure*}

We train PS-DLO on self-collected demonstrations and evaluate it on nine real-world DLO manipulation tasks, learning a shared multi-task policy separately for the bimanual and single-arm platforms. PS-DLO achieves an average success rate of 68.9\%, exceeding the strongest baseline by 20.0 percentage points and achieving the highest success rate on eight of the nine tasks. Controlled ablations show that the full framework outperforms either component alone. We further evaluate butterfly-knot untying under seven conditions involving individual and combined changes in rope color, lighting, and background. PS-DLO outperforms baseline in six conditions and matches its performance in the remaining one. When all three factors change simultaneously, PS-DLO achieves a success rate of 10.0\%, compared with 0.0\% for baseline, indicating improved performance under these combined visual changes.The code for PS-DLO is publicly available at
\url{https://github.com/LMD4790/PS-DLO/}.

Our main contributions are threefold:
\begin{itemize}
    \item We introduce object-property conditioning for multi-task DLO manipulation, using categorical descriptions inferred from initial observations to provide explicit physical property for action generation.

    \item We propose mask-guided future representation distillation, using an action-supervised teacher--assistant framework to distill future DLO spatial knowledge into the policy's action expert.

    \item We evaluate PS-DLO on nine real-world DLO tasks across single-arm and bimanual platforms, achieving an average success rate of 68.9\%, exceeding $\pi_{0.5}$ by 20.0 percentage points.
\end{itemize}

\section{Related Work}

\subsection{DLO Manipulation}

Existing DLO manipulation methods use dense descriptors and goal-conditioned Transporter Networks for object correspondence and rearrangement~\cite{sundaresan2020dense,seita2021gctn}. Other approaches exploit deformation models and feedback control~\cite{wang2022offlineonline,yu2023coarse}, tactile motion primitives~\cite{wilson2023cable}, or cable tracing and crossing classification~\cite{viswanath2021disentangling,viswanath2023handloom} to support task-specific manipulation~\cite{yao2026manipulation}.
Recent work considers property adaptation and broader task coverage. Li and Choi condition insertion on DLO flexibility estimated after a predefined grasping interaction~\cite{li2024flexibility}, while Hierarchical DLO Routing combines VLM-based planning with learned and predefined skills~\cite{li2026hierarchical}. DLO-Lab and WireCraft provide simulation environments spanning multiple DLO tasks~\cite{cao2026dlolab,wirecraft2026}, with DLO-Lab also demonstrating joint multi-task VLA training. Building on these directions, our work investigates mask-guided representation supervision and categorical physical-semantic conditioning for real-world multi-task DLO manipulation.

\subsection{Representation Learning for VLA Policies}

VLA models adapt pretrained vision-language representations
for robotic control~\cite{rdt, zhang2025large},
with different interfaces for incorporating proprioception.
$\pi_0$ projects continuous states into action-expert tokens,
whereas $\pi_{0.5}$ discretizes them into its textual input
prefix~\cite{black2025pi0,black2025pi05}.
Beyond input conditioning, SAM3D-guided alignment and AffordVLA
introduce object-level geometry and task-conditioned affordances
through visual representation
supervision~\cite{liu2026sam3dvla,affordvla2026}.

Predictive representations further connect visual learning
with the consequences of robot actions. Mask World Model
forecasts future semantic masks and uses mask-centric features
for control, reporting improved robustness to appearance
changes~\cite{lou2026mwm}. Efficient-WAM reduces the cost of
future visual prediction through video-expert distillation
while retaining the prediction branch during
deployment~\cite{li2026efficientwam}. FLARE incorporates future
information through alignment between intermediate policy
features and action-aware future embeddings, without computing
target embeddings at inference~\cite{zheng2025flare}.
Building on these advances, our work combines SAM3-selected
next-frame supervision within the action expert with CLIP-derived
DLO property descriptions supplied through the existing
VLA text-conditioning interface~\cite{radford2021clip}.

\section{Method}

\subsection{Problem Formulation}
\label{sec:problem_formulation}

We consider multi-task manipulation of DLO with a shared policy for each robot embodiment.
At time $t$, the inputs are RGB observations
$\mathbf{O}_t=\{O_t^v\}_{v=1}^{V}$, a language task instruction
$l$, and the robot's proprioceptive state $s_t$, where $O_t^v$ denotes the image from view $v$, and $V$ is the
number of camera views used in our experiments. A policy $\pi_{\theta}$,
parameterized by $\theta$, predicts an action chunk:

\begin{equation}
\widehat{\mathbf{A}}_t \sim \pi_\theta(\cdot \mid\mathbf{O}_t, l, s_t),
\qquad
\widehat{\mathbf{A}}_t \in \mathbb{R}^{H \times D_a},
\end{equation}
where $\widehat{\mathbf{A}}_t$ contains $H$ consecutive predicted
actions starting at $t$, and $D_a$ denotes the model-space action
dimension. Our objective is to learn, from demonstrations, a shared
vision-language-conditioned policy for each robot embodiment that
performs diverse DLO manipulation tasks across variations
in object properties and configurations.


\subsection{Framework Overview}




In DLO manipulation, interaction regions are difficult to identify, and differences in object properties affect action selection. We use physical-semantic conditioning and local visual supervision to support action learning. As shown in Fig.~\ref{fig:architecture}, PS-DLO builds on a flow-matching VLA model backbone~\cite{black2025pi05}, incorporates the extracted property prompt through the existing language input pathway, and adds learnable DLO queries and a representation alignment head to the action expert. Training consists of three stages: cross-modal adaptation, teacher pretraining, and joint training.

\textbf{Stage 1: Cross-modal adaptation.} We construct an extractor based on contrastive image-text learning~\cite{radford2021clip} to align representations of initial main-view images and DLO-property prompts. The extractor is trained with the InfoNCE contrastive loss $\mathcal{L}_{\mathrm{contrast}}$.
This stage grounds property descriptions in DLO observations,
enabling the extractor to provide object-specific conditioning
for action learning.
After adaptation, we freeze the extractor throughout teacher
pretraining and joint training.

\textbf{Stage 2: Teacher pretraining.} A frozen SAM3 model extracts visual features and a DLO mask from the next-frame main-view image using a concept prompt. Mask-guided patch selection retains the relevant regions. The task prompt and the property prompt are encoded by the SAM3 text encoder and passed to the Teacher Encoder together with the selected visual features. A fusion module and a Q-former produce teacher tokens, which condition an auxiliary action Diffusion Transformer (DiT) through cross-attention. The auxiliary action DiT also receives the proprioceptive state and noisy action chunk. We jointly optimize the Teacher Encoder and auxiliary action DiT using a flow-matching loss $\mathcal{L}_{\mathrm{aux}}$.

\textbf{Stage 3: Joint training.} The TA encoder is initialized from the pretrained Teacher Encoder's parameters. During joint training, the TA encoder and auxiliary action DiT continue to receive flow-matching supervision, while the Teacher Encoder is updated only by EMA. The student policy learns to generate action chunks using a flow-matching loss on demonstration actions. Its intermediate DLO-query representations are projected through the alignment head, and their mean cosine distance to the stop-gradient Teacher query representations serves as a future constraint on action learning.

\subsection{DLO Property Extraction}

To provide object-property conditioning in addition to the task instruction, we build a DLO property extractor that maps the initial observation to an object-level semantic description through image--text similarity matching. The property prompt dataset is denoted by $\mathcal D=\{d_c\}_{c=1}^{C}$, where $C$ is the number of prompts and each $d_c$ contains information on object count, length, thickness, stiffness, extensibility and initial configuration.

The image and text encoders map the initial third-person view and the property prompt to features $z_0$ and $e_c$, respectively. We apply $L_2$ normalization and compute temperature-scaled cosine similarity scores:
\begin{equation}
\ell_c
=
\exp(\alpha)
\frac{z_0^\top e_c}{\|z_0\|_2\|e_c\|_2},
\end{equation}
where $\ell_c$ is the matching score and $\alpha$ is the model's stored logit-scale parameter. We then apply softmax to obtain normalized matching probabilities $p_c$ and select the highest-scoring description:
\begin{equation}
p_c=\frac{\exp(\ell_c)}{\sum_{j=1}^{C}\exp(\ell_j)},
\qquad
d_0=d_{\arg\max_{c}p_c},
\end{equation}
where $j$ indexes the candidate descriptions and $d_0$ denotes the output description.

The description $d_0$ is computed once at the start of each episode and remains fixed throughout the episode. We serialize $d_0$, the task instruction $l$, and the discretized proprioceptive state $s_t$ in that order and pass the resulting sequence through the existing language input pathway of the PS-DLO backbone. Together with the current RGB observations from $V$ views, this sequence conditions action generation. The description provides a category-level prior over the initial object properties and configuration; it does not serve as a measurement of continuous physical parameters or a dynamic state estimate.
\subsection{Spatial Knowledge Distillation}

To make the action expert spatially aware, we introduce a teacher--assistant spatial knowledge distillation scheme that leverages future DLO observations available during training to provide explicit spatial supervision. Given a third-person view $O_{t+1}^{\mathrm{main}}$,
a frozen SAM3 model extracts visual features and a concept-prompted
DLO mask. We retain mask-covered patches, preserve their spatial
positions through positional embeddings, and concatenate them with
SAM3 text features of the task instruction $l$ and the property description $d_0$ produced by the DLO property extractor. These inputs, denoted by $X_t$, pass through four fusion Transformer blocks and a two-block Q-former with $Q$
learnable queries, producing $256$-dimensional tokens.
We first pretrain the Teacher Encoder $E_{\theta_T}$ together with
a randomly initialized, eight-block conditional action DiT for
$20{,}000$ updates on the demonstration training split.
Conditioned on the encoder tokens and current state $s_t$, the DiT
predicts the flow velocity of a noisy demonstration action chunk.

For $A_t^\tau=(1-\tau)A_t+\tau\epsilon$, where
$\epsilon\sim\mathcal{N}(0,I)$, the auxiliary objective is
\begin{equation}
\mathcal{L}_{\mathrm{aux}}
=
\mathbb{E}\!\left[
\left\|
v_{\psi}\!\left(A_t^\tau,\tau,s_t,E_{\theta_T}(X_t)\right)
-(\epsilon-A_t)
\right\|_F^2
\right].
\label{eq:future_flow}
\end{equation}

At the start of joint training, we copy the pretrained Teacher Encoder
into a TA, initializing
$\theta_{\mathrm{TA}}\leftarrow\theta_T$.
The TA retains the pretrained auxiliary action DiT.
During joint training, the TA encoder and auxiliary action DiT
are jointly optimized using $\mathcal{L}_{\mathrm{aux}}$,
with $E_{\theta_{\mathrm{TA}}}$ replacing $E_{\theta_T}$
in Eq.~\eqref{eq:future_flow}.
The Teacher independently processes the same $X_t$ without
backpropagation and follows the TA through EMA,
$\theta_T\leftarrow m\theta_T+(1-m)\theta_{\mathrm{TA}}$,
with $m=0.995$, providing temporally smoothed distillation targets.
In the student, we append $Q$ learnable DLO queries to the action
tokens and project their layer-14 hidden states $h_{t,q}^{S}$ into
the teacher feature space using an MLP $g$.
Writing $z_{t,q}^{T}$ for the teacher's $q$ th output token, we minimize
\begin{equation}
\mathcal{L}_{\mathrm{future}}
=
\mathbb{E}_{t\in\mathcal{V}}
\left[
\frac{1}{Q}\sum_{q=1}^{Q}
\left(
1-\operatorname{cos}\!\left(
g(h_{t,q}^{S}),\operatorname{sg}(z_{t,q}^{T})
\right)
\right)
\right],
\label{eq:spatial_distillation}
\end{equation}
where $\operatorname{sg}$ denotes stop-gradient and $\mathcal{V}$
contains samples with an available future frame and a valid DLO mask.
The student optimizes
$\mathcal{L}_{\mathrm{action}}
+\lambda \cdot\mathcal{L}_{\mathrm{future}}$, where $\lambda =0.1$,
while the TA and auxiliary DiT receive gradients only from
$\mathcal{L}_{\mathrm{aux}}$.
Samples outside $\mathcal{V}$ retain student action supervision.

\begin{figure*}[t]
    \centering
    \includegraphics[width=0.98\textwidth]{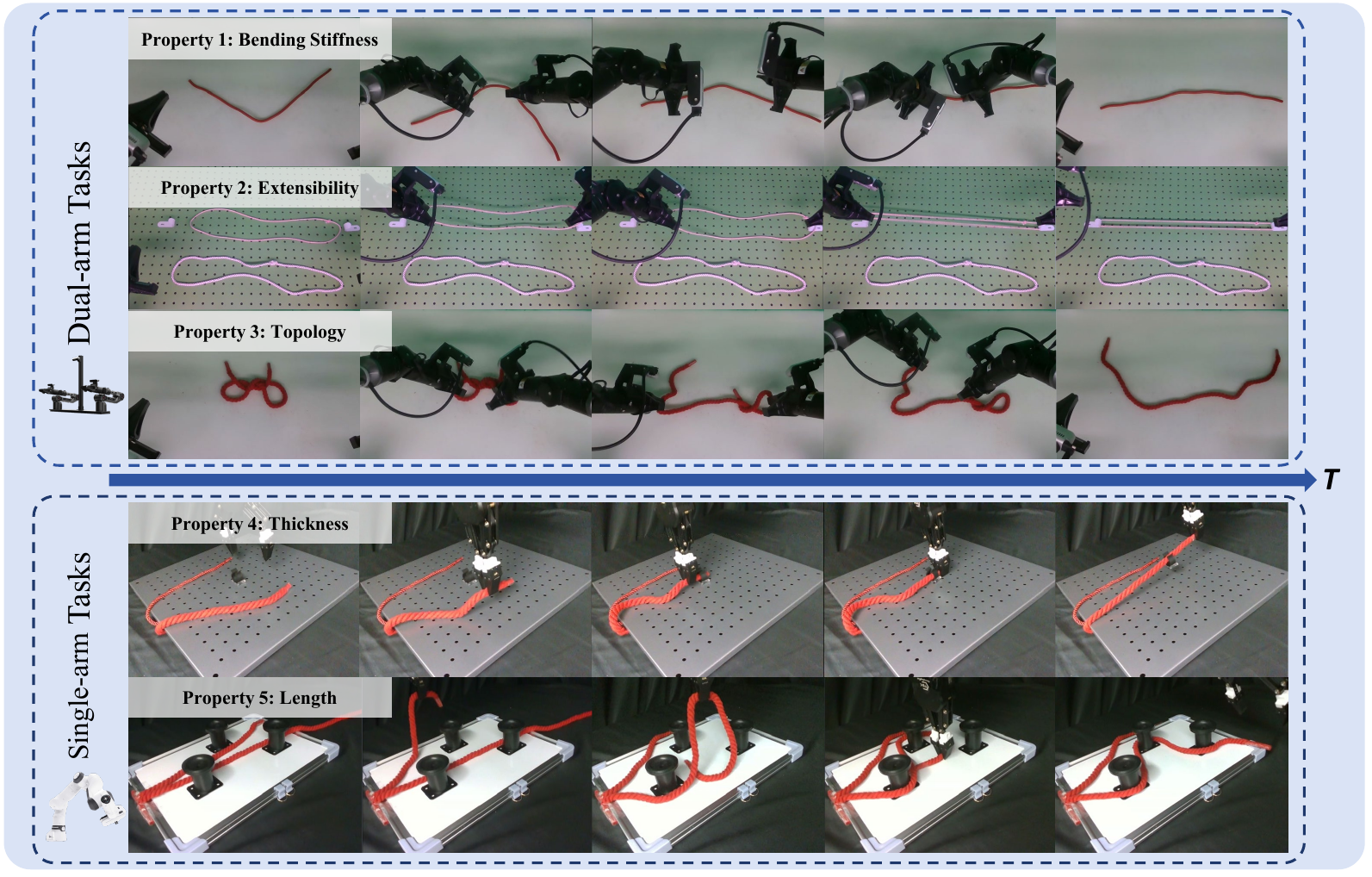}
    \caption{
    \textbf{Representative execution sequences of real-world
    DLO manipulation tasks.}
    The upper three rows show straightening, elastic-rope hooking,
    and knot untying on the bimanual PiPER platform.
    The lower two rows show  thick-rope threading and lifting and long-rope wrapping on the single-arm Franka Research 3 platform.
    The tasks cover five DLO property dimensions:
    bending stiffness, extensibility, topology, thickness, and length.
    Frames are arranged chronologically from left to right.
    }
    \label{fig:task settings}
\end{figure*}

\section{Experiments}
\label{sec:experiments}

\subsection{Experiment Setup}

We conduct experiments on two robot embodiments: a dual-arm PiPER and a single-arm Franka Research 3. For each embodiment, we train a shared policy across all of its tasks. We define five task categories around bending stiffness, topology, extensibility, length, and thickness. PiPER performs three categories: bending stiffness, topology, and extensibility.
Bending stiffness involves cable straightening and rope straightening, while topology
covers single-knot, bow-tie, butterfly-knot, and figure-eight-knot untying. Extensibility is evaluated through elastic-rope hooking. Franka Research 3 performs the remaining two categories: length (long rope wrapping) and thickness (thick-rope threading and lifting), as shown in Fig.~\ref{fig:task settings}. In addition to the five task-defining property dimensions, we construct a property-prompt dataset that describes each demonstration using six categorical attributes---the number of DLO, bending stiffness, topology, extensibility, length, and thickness---for property extraction and policy conditioning.

\subsection{Data and Evaluation Protocol}

Each demonstration contains multi-view RGB observations, proprioceptive state, action sequences, and the corresponding task and property prompts. After cleaning, the dataset contains 500 demonstrations: 100 for straightening (50 cable and 50 rope), 100 for knot untying (25 each for single-knot, bow-tie, butterfly, and figure-eight knots), and 100 each for elastic-rope hooking, long-rope wrapping, and thick-rope threading followed by lifting.

For a fair comparison, all baseline models use exactly the same training data as PS-DLO and are fully fine-tuned for 10,000 optimization steps. During training, performance is monitored on the validation set, and the checkpoint with the lowest validation loss is selected for real-robot evaluation.

Each trial has a maximum duration of 60\,s, with additional task-specific success criteria. Cable straightening is considered successful when the resulting cable configuration has an opening angle greater than $60^\circ$, whereas rope straightening requires that no bend exceeds $90^\circ$. A knot-untying trial succeeds only when the knot is completely removed. Elastic-rope hooking requires the rope to remain securely engaged with the hook after both grippers are released. Long-rope wrapping succeeds when the final rope configuration forms the specified S-shaped path around the three posts. For thick-rope threading and lifting, the rope must pass through the target hole and remain stably lifted for 10\,s after threading.

\begin{table*}[htbp]
\centering
\caption{SUCCESS RATES (\%) ON NINE DLO MANIPULATION TASKS SPANNING FIVE DLO PROPERTIES. ALL METHODS ARE EVALUATED UNDER THE SAME REAL-ROBOT PROTOCOL, WITH 20 TRIALS PER TASK.}
\label{tab:main_results}
\scriptsize
\setlength{\tabcolsep}{2.6pt}
\renewcommand{\arraystretch}{1.08}
\resizebox{\textwidth}{!}{%
\begin{tabular}{ll*{9}{c}}
\toprule
\textbf{Property}
& \textbf{Task}
& \textbf{OpenVLA}~\cite{kim2024openvla}
& $\boldsymbol{\pi_{0}}$~\cite{black2025pi0}
& $\boldsymbol{\pi_{0}}$-\textbf{FAST}~\cite{fast}
& $\boldsymbol{\pi_{0.5}}$~\cite{black2025pi05}
& \textbf{OpenVLA-OFT}~\cite{openvla-oft}
& \textbf{LingBotVLA 2.0}~\cite{LingBotVLA2}
& \textbf{GR00T N1.7}~\cite{n1.7}
& \textbf{DreamZero}~\cite{dreamzero}
& \textbf{Ours} \\
\midrule

Bending stiffness
& Cable straightening
& 15.0 & 25.0 & 20.0 & 35.0 & 30.0
& 45.0 & 20.0 & 30.0 & \textbf{60.0} \\

Bending stiffness
& Rope straightening
& 25.0 & 40.0 & 30.0 & 45.0 & 45.0
& 45.0 & 30.0 & 40.0 & \textbf{80.0} \\

Topology
& Single-knot untying
& 25.0 & 40.0 & 30.0 & 50.0 & 45.0
& 50.0 & 35.0 & 45.0 & \textbf{75.0} \\

Topology
& Bow-tie untying
& 25.0 & 40.0 & 30.0 & 50.0 & 45.0
& 45.0 & 35.0 & 45.0 & \textbf{70.0} \\

Topology
& Butterfly-knot untying
& 35.0 & 60.0 & 40.0 & 70.0 & 65.0
& 70.0 & 55.0 & 65.0 & \textbf{90.0} \\

Topology
& Figure-eight-knot untying
& 30.0 & 50.0 & 35.0 & 60.0 & 55.0
& 60.0 & 45.0 & 55.0 & \textbf{85.0} \\

Extensibility
& Elastic-rope hooking
& 10.0 & 20.0 & 15.0 & 25.0 & 25.0
& 20.0 & 15.0 & \textbf{30.0} & 20.0 \\

Length
& Long-rope wrapping
& 35.0 & 60.0 & 40.0 & 75.0 & 70.0
& 65.0 & 55.0 & 70.0 & \textbf{80.0} \\

Thickness
& Thick-rope threading and lifting
& 15.0 & 25.0 & 20.0 & 30.0 & 30.0
& 30.0 & 20.0 & 30.0 & \textbf{60.0} \\

\midrule
\multicolumn{2}{l}{\textbf{Overall average}}
& 23.9 & 40.0 & 28.9 & 48.9 & 45.6
& 47.8 & 34.4 & 45.6 & \textbf{68.9} \\
\bottomrule
\end{tabular}%
}
\end{table*}

\subsection{Main Results and Analysis}

\textbf{Attribute recognition.}
We evaluate the DLO property extractor on a held-out test set constructed from 50 separately collected trajectories that were excluded from model training. From the first three seconds of each trajectory, we randomly sample 10 frames within each one-second interval, yielding 1,500 candidate frames. After manually removing visually duplicate frames, we obtain 500 test images used exclusively for evaluation. A prediction is considered correct only if all predicted attributes match the ground-truth annotation. The extractor achieves an accuracy of $99.4\%$ on this test set. This metric measures the recognition of categorical DLO properties rather than the accuracy of continuous physical parameter estimates.

\textbf{Overall comparison.}
Table~\ref{tab:main_results} summarizes success rates on nine tasks spanning five task categories. Each reported method--task pair is evaluated over $20$ real-robot trials, yielding $180$ trials per method. Overall performance is calculated as the macro-average of task success rates. PS-DLO achieves an average success rate of $68.9\%$ and the highest success rate on eight of the nine tasks. Among baselines evaluated on all tasks, $\pi_{0.5}$ achieves the highest average success rate of $48.9\%$. PS-DLO exceeds this baseline by $20.0$ percentage points and outperforms LingBotVLA 2.0, OpenVLA-OFT, and DreamZero by $21.1$, $23.3$, and $23.3$ percentage points, respectively.

\textbf{Straightening, knot-untying and elastic-rope hooking.}
On the dual-arm PiPER, PS-DLO achieves success rates of $60.0\%$ and $80.0\%$ on cable and rope straightening, exceeding $\pi_{0.5}$ by $25.0$ and $35.0$ percentage points, respectively. Its success rates on single-knot, bow-tie, butterfly-knot, and figure-eight-knot untying are $75.0\%$, $70.0\%$, $90.0\%$, and $85.0\%$, respectively. The average across these four tasks is $80.0\%$, compared with $57.5\%$ for $\pi_{0.5}$. PS-DLO thus achieves higher success rates than the baselines for both straightening objects and all four knot types evaluated. Elastic-rope hooking is the only task on which PS-DLO does not achieve the highest success rate. Its success rate is $20.0\%$, below the $25.0\%$ of $\pi_{0.5}$ and the $30.0\%$ of DreamZero. The gains are therefore not uniform across tasks, with no improvement over $\pi_{0.5}$ on elastic-rope hooking. The limitations on this task are discussed further
in Section~\ref{sec:discussion}.

\textbf{Wrapping and threading tasks.}
On the single-arm Franka, PS-DLO achieves success rates of $80.0\%$ on long-rope wrapping and $60.0\%$ on thick-rope threading and lifting, exceeding $\pi_{0.5}$ by $5.0$ and $30.0$ percentage points, respectively. The improvement is relatively small on long-rope wrapping, where $\pi_{0.5}$ already achieves a success rate of $75.0\%$.

\begin{figure}[t]
    \centering
    \includegraphics[width=\columnwidth]{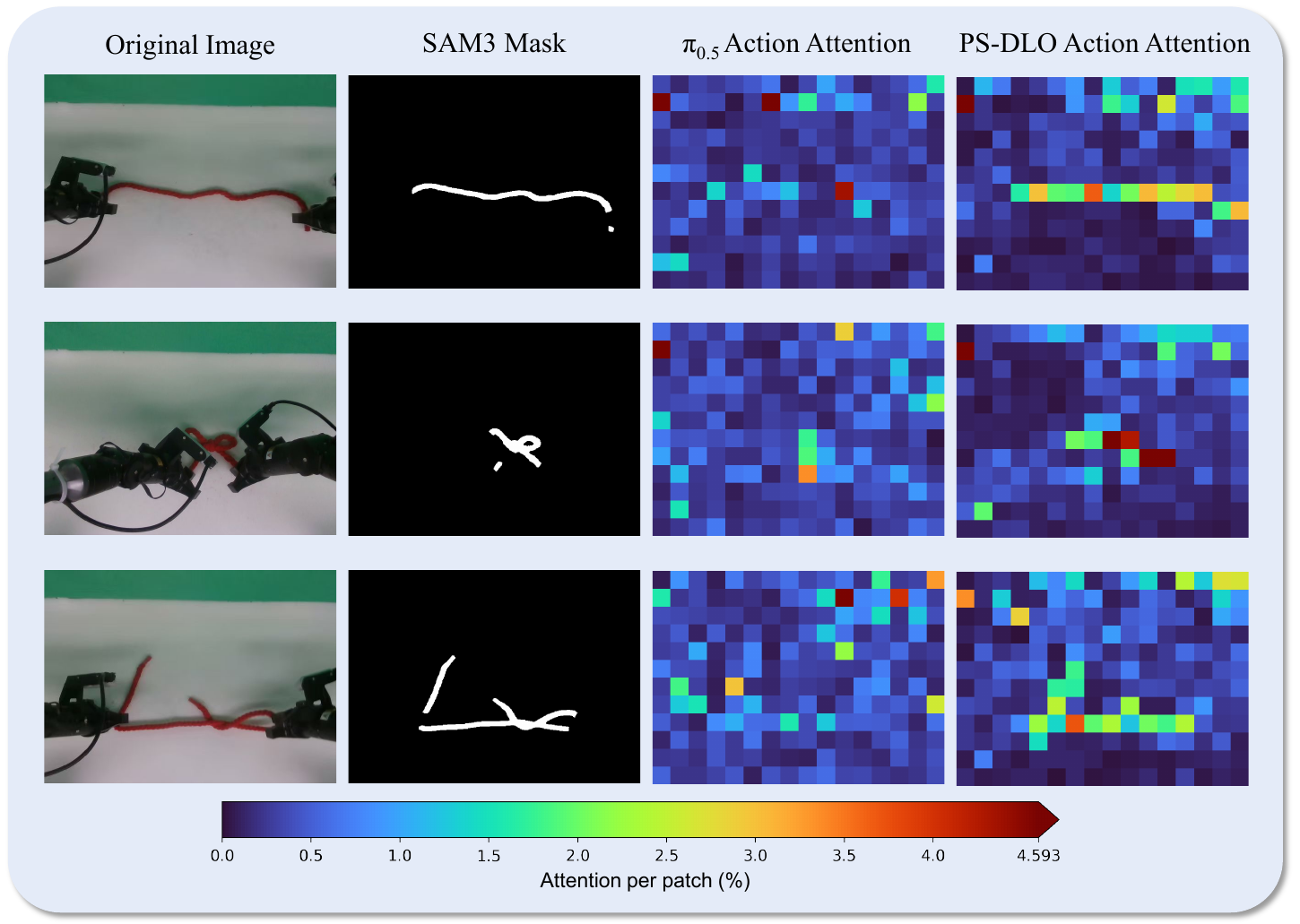}
    \caption{
    \textbf{Qualitative comparison of action attention distributions.}
    In the displayed examples, PS-DLO exhibits more concentrated
    attention around DLO structures and knot regions, whereas
    $\pi_{0.5}$ shows more dispersed responses.
    }
    \label{fig:heat map}
\end{figure}

\textbf{Action attention visualization.}
Fig.~\ref{fig:heat map} compares the action attention distributions of
PS-DLO and $\pi_{0.5}$ under the same observations, using SAM3 masks
as references for cable locations. In the examples shown,
$\pi_{0.5}$ exhibits more scattered high-response regions, whereas
those of PS-DLO are more concentrated around the cables and knots.

\begin{table}[t]
\centering
\caption{
Ablation results grouped by property conditioning,
spatial supervision, and teacher-target updates.
SR denotes success rate.
Peak training memory is reported per GPU.
}
\label{tab:configuration_memory}
\footnotesize
\setlength{\tabcolsep}{3pt}
\renewcommand{\arraystretch}{1.15}
\sisetup{
    mode = text,
    detect-weight = true
}

\begin{tabular*}{\linewidth}{
    @{\extracolsep{\fill}}
    l
    S[table-format=2.1]
    S[table-format=2.2]
    S[table-format=1.3]
    @{}
}
\toprule
\multirow{2}{*}{\textbf{Configuration}}
& \multicolumn{1}{c}{
    \multirow{2}{*}{\textbf{SR (\%) $\uparrow$}}
}
& \multicolumn{2}{c@{}}{
    \textbf{Peak memory (GB) $\downarrow$}
} \\
\cmidrule(l){3-4}
& & {\textbf{Train}} & {\textbf{Infer.}} \\
\midrule

Baseline ($\pi_{0.5}$)
& 48.9 & 15.96 & 8.263 \\

\midrule
\multicolumn{4}{l}{
    \textbf{A. Property fusion strategy}
} \\
\addlinespace[2pt]

Prefix conditioning
& 50.0 & 21.63 & 8.634 \\

Suffix conditioning
& 56.1 & 16.90 & 8.625 \\

Joint text conditioning
& 49.4 & 16.86 & 8.617 \\

\midrule
\multicolumn{4}{l}{
    \textbf{B. SAM3-based spatial supervision}
} \\
\addlinespace[2pt]

Future-frame SAM3
& 57.2 & 16.86 & 8.263 \\

\begin{tabular}[c]{@{}l@{}}
Joint text conditioning \\
+ current-frame SAM3
\end{tabular}
& 61.1 & 16.86 & 8.617 \\

\midrule
\multicolumn{4}{l}{
    \textbf{C. Teacher-target update strategy}
} \\
\addlinespace[2pt]

Fixed pretrained teacher
& 63.3 & 16.01 & 8.617 \\

\textbf{TA + EMA teacher (Ours)}
& \bfseries 68.9 & 16.86 & 8.617 \\

\bottomrule
\end{tabular*}
\end{table}

\subsection{Ablation Study}

Using $\pi_{0.5}$ as the baseline, we organize the ablation
experiments into three groups: property fusion strategies,
SAM3-based spatial supervision, and teacher-target updates.
Table~\ref{tab:configuration_memory} reports the success rate,
peak training memory per GPU, and peak inference memory
of each configuration.

\textbf{Property fusion strategy.}
We compare prefix conditioning, suffix conditioning, and
joint text conditioning for incorporating DLO property information.
Joint text conditioning combines the task instruction,
discretized robot state, and DLO property description
within the policy's text input.
The three variants achieve success rates of $50.0\%$, $56.1\%$,
and $49.4\%$, respectively, compared with $48.9\%$ for the baseline.
Suffix conditioning performs best among the variants using
property conditioning alone, improving success by $7.2$
percentage points.
These results indicate that the effectiveness of property
conditioning depends on its integration into the policy.

\textbf{SAM3-based spatial supervision.}
The future-frame SAM3 configuration achieves a success rate
of $57.2\%$, exceeding the baseline by $8.3$ percentage points.
Combining joint text conditioning with SAM3-based supervision
from the current observation achieves $61.1\%$, compared with
$49.4\%$ for joint text conditioning alone, an improvement of
$11.7$ percentage points.
These comparisons support the usefulness of mask-guided
representation supervision in the evaluated configurations.
The current-frame variant constructs its targets from $O_t$,
whereas the full method uses the next recorded observation
$O_{t+1}$.

\textbf{Teacher-target update strategy.}
Using a fixed pretrained Teacher Encoder to provide alignment
targets yields a success rate of $63.3\%$.
In the full method, the TA encoder and auxiliary action DiT
are optimized through auxiliary action supervision, while
the Teacher Encoder is updated through an exponential moving
average of the TA encoder's parameters.
The full method achieves $68.9\%$, exceeding the fixed-teacher
variant by $5.6$ percentage points.
This comparison evaluates the combined use of online TA
optimization and EMA teacher updates; it does not isolate
the effect of EMA smoothing.

\subsection{Generalization Study}

\begin{figure}[t]
    \centering
    \includegraphics[width=\columnwidth]{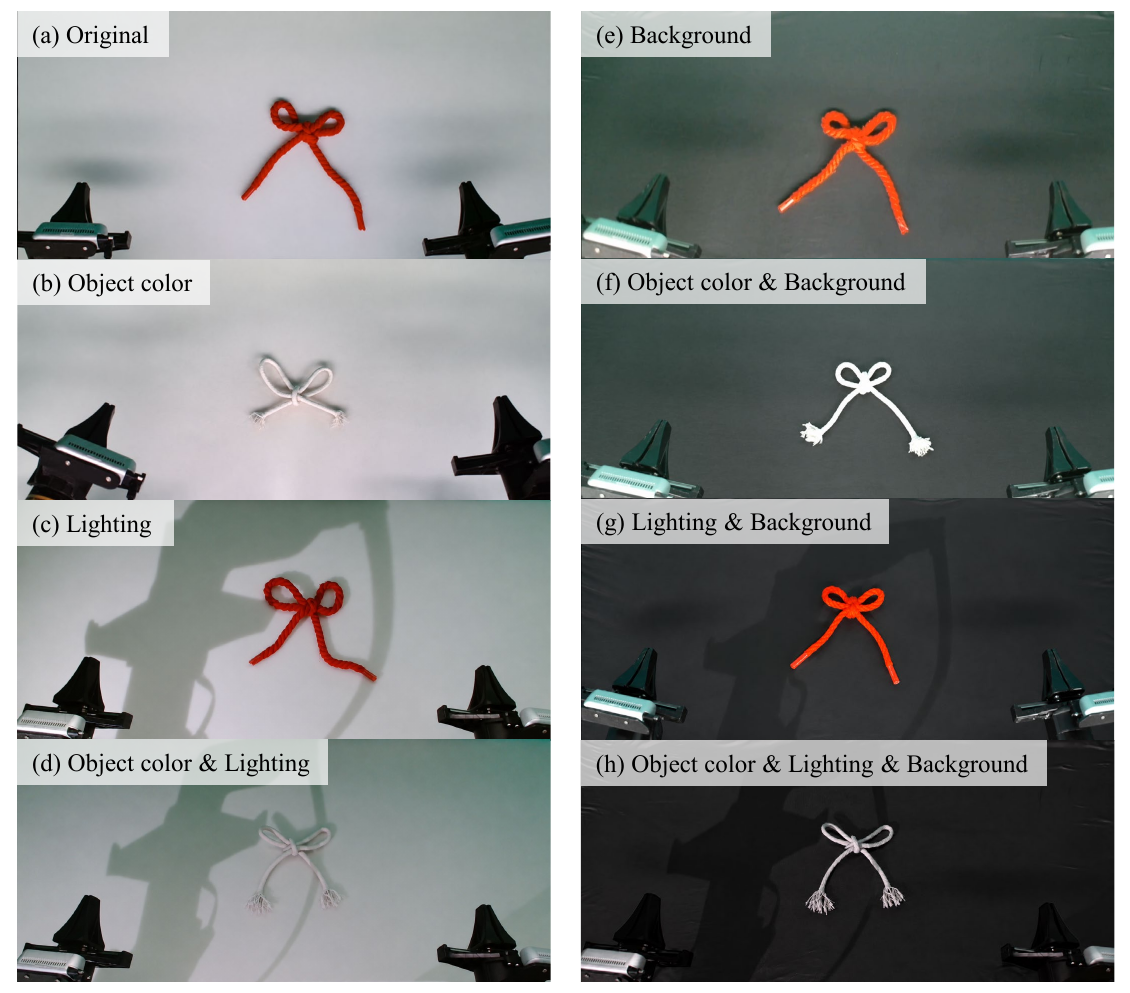} 
    \caption{
\textbf{Visual conditions for the generalization study.}
Representative observations of the butterfly-knot untying task
under eight conditions:
(a) original setting;
(b) object color change;
(c) lighting change;
(d) combined color and lighting changes;
(e) background change;
(f) combined color and background changes;
(g) combined lighting and background changes; and
(h) simultaneous color, lighting, and background changes.
}
\label{fig:generalization}
\end{figure}

We evaluate robustness to visual changes on the butterfly-knot
untying task by varying object color, lighting, and background,
as illustrated in Fig.~\ref{fig:generalization}.
For the color change, we replace the red rope with a white rope
of the same material while retaining the knot type.
For the lighting change, we introduce random illumination
perturbations.
For the background change, we cover the original white surface
with a thin black cloth.
These changes are applied individually and in all pairwise
and three-factor combinations, yielding seven shifted conditions
in addition to the original setting.
Each method is evaluated over 20 trials per condition.

Table~\ref{tab:generalization} summarizes the results.
PS-DLO achieves higher success rates than $\pi_{0.5}$ in six
of the seven shifted conditions and matches its performance
under the background-only change.
Under individual color, lighting, and background changes,
PS-DLO achieves success rates of $80.0\%$, $85.0\%$, and
$70.0\%$, respectively.
Under combined color and lighting, color and background,
and lighting and background changes, PS-DLO achieves
$65.0\%$, $40.0\%$, and $45.0\%$, exceeding $\pi_{0.5}$
by $20.0$, $15.0$, and $15.0$ percentage points, respectively.
When all three changes are applied simultaneously,
PS-DLO succeeds in 2 of 20 trials ($10.0\%$), while
$\pi_{0.5}$ succeeds in none.
These results suggest improved robustness across the evaluated
conditions on this task, although the low success rates under
the three-factor condition reveal a limitation under combined
visual changes.

\section{Discussion}
\label{sec:discussion}

\begin{table}[t]
\centering
\caption{
Success rates (\%) on butterfly-knot untying under
visual shifts, with 20 trials per method per condition.
}
\label{tab:generalization}
\scriptsize
\setlength{\tabcolsep}{4pt}
\renewcommand{\arraystretch}{1.08}

\begin{tabular*}{\linewidth}{
    @{\extracolsep{\fill}}lcc@{}
}
\toprule
\textbf{Visual condition}
& $\boldsymbol{\pi_{0.5}}$
& \textbf{Ours} \\
\midrule

Original
& 70.0 & \textbf{90.0} \\

\addlinespace[3pt]
Object color
& 60.0 & \textbf{80.0} \\

Lighting
& 70.0 & \textbf{85.0} \\

Background
& \textbf{70.0} & \textbf{70.0} \\

\addlinespace[3pt]
Object color + lighting
& 45.0 & \textbf{65.0} \\

Object color + background
& 25.0 & \textbf{40.0} \\

Lighting + background
& 30.0 & \textbf{45.0} \\

\addlinespace[3pt]
Object color + lighting + background
& 0.0 & \textbf{10.0} \\

\midrule
\textbf{Average (all 8 conditions)}
& 46.3 & \textbf{60.6} \\
\bottomrule
\end{tabular*}
\end{table}

\subsection{Teacher--Assistant Design}

Updating the teacher through the action-supervised TA improves the average success rate from 63.3\% to 68.9\% compared with keeping the pretrained teacher fixed (Table~\ref{tab:configuration_memory}). This suggests that teacher pretraining alone does not fully capture the representations needed for the evaluated manipulation tasks. The TA continues to refine these representations through action prediction, while the teacher follows its parameters through EMA, drawing on the weight-averaging principle used in Mean Teacher~\cite{tarvainen2017mean}. This design allows the supervision to evolve during policy training while smoothing teacher updates. The observed gain supports the combined design; isolating the benefit of EMA would require a further comparison using the TA outputs directly as alignment targets.

\subsection{Interpretation of Generalization Results}

PS-DLO outperforms $\pi_{0.5}$ in six of the seven visual-shift conditions on butterfly-knot untying (Table~\ref{tab:generalization}), with gains of 15.0--20.0 percentage points under pairwise combinations of color, lighting, and background changes. One possible explanation is that SAM3-based spatial supervision encourages the policy to retain DLO-relevant features despite changes in scene appearance. This interpretation is consistent with prior work showing that object-centric representations can improve manipulation generalization under visual changes~\cite{zhu2023learning}. However, the advantage narrows under more challenging conditions: when all three factors change, PS-DLO succeeds in only 2 of 20 trials. Higher success rates also do not imply smaller performance drops; under the background-only change, PS-DLO falls from 90.0\% to 70.0\%, while $\pi_{0.5}$ remains at 70.0\%. These results show that PS-DLO retains a performance advantage under most tested shifts, but remains sensitive to visual variation.

\subsection{Limitations on Elastic-Rope Hooking}

Elastic-rope hooking is the only task where PS-DLO does not achieve the highest success rate, reaching 20.0\% compared with DreamZero's 30.0\% (Table~\ref{tab:main_results}). Beyond rope--hook positioning, this task requires handling elastic recoil and maintaining hook engagement after release. DreamZero jointly predicts future video and actions~\cite{dreamzero}, which may help anticipate these dynamic changes. PS-DLO instead uses future observations for representation supervision during training, while its episode-level property description provides a static prior. This performance gap motivates extending spatial supervision toward explicit prediction of deformation and contact transitions, although the current comparison does not isolate the effect of future prediction.

\section{Conclusion}
We presented PS-DLO, which combines object-property conditioning
with future spatial representation distillation through an
action-supervised TA and an EMA-updated teacher.
Across nine real-world DLO manipulation tasks, PS-DLO achieves
an average success rate of $68.9\%$, exceeding the best baseline by $20.0$ percentage points.
It also achieves higher success rates under most tested
visual shifts, although elastic-rope hooking and combined
visual changes remain challenging.
Future work will investigate these failure modes and extend
evaluation to more diverse tasks and object properties.

\bibliographystyle{IEEEtran}
\bibliography{ref}

@IEEEtranBSTCTL{IEEEBSTcontrol,
  CTLuse_forced_etal       = "yes",
  CTLmax_names_forced_etal = "6",
  CTLnames_show_etal       = "1",
  CTLdash_repeated_names = "no"
}

@article{caporali2026survey,
  author  = {Alessio Caporali and Ignacio Cuiral-Zueco and Gonzalo L{\'o}pez-Nicol{\'a}s and Gianluca Palli},
  title   = {Robotic Perception and Manipulation of Deformable Linear Objects: A Survey},
  journal = {The International Journal of Robotics Research},
  year    = {2026},
  month   = mar,
  doi     = {10.1177/02783649261432253},
  url     = {https://doi.org/10.1177/02783649261432253}
}

@inproceedings{li2024flexibility,
  author    = {Mingen Li and Changhyun Choi},
  title     = {Learning for Deformable Linear Object Insertion Leveraging Flexibility Estimation from Visual Cues},
  booktitle = {2024 IEEE International Conference on Robotics and Automation (ICRA)},
  year      = {2024},
  pages     = {5183--5189},
  doi       = {10.1109/ICRA57147.2024.10610419},
  url       = {https://doi.org/10.1109/ICRA57147.2024.10610419}
}

@article{wang2022offlineonline,
  author  = {Changhao Wang and Yuyou Zhang and Xiang Zhang and Zheng Wu and Xinghao Zhu and Shiyu Jin and Te Tang and Masayoshi Tomizuka},
  title   = {Offline-Online Learning of Deformation Model for Cable Manipulation With Graph Neural Networks},
  journal = {IEEE Robotics and Automation Letters},
  year    = {2022},
  volume  = {7},
  number  = {2},
  pages   = {5544--5551},
  month   = apr,
  doi     = {10.1109/LRA.2022.3158376},
  url     = {https://doi.org/10.1109/LRA.2022.3158376}
}

@inproceedings{yu2023coarse,
  author    = {Mingrui Yu and Kangchen Lv and Changhao Wang and Masayoshi Tomizuka and Xiang Li},
  title     = {A Coarse-to-Fine Framework for Dual-Arm Manipulation of Deformable Linear Objects with Whole-Body Obstacle Avoidance},
  booktitle = {2023 IEEE International Conference on Robotics and Automation (ICRA)},
  year      = {2023},
  pages     = {10153--10159},
  doi       = {10.1109/ICRA48891.2023.10160264},
  url       = {https://doi.org/10.1109/ICRA48891.2023.10160264}
}

@inproceedings{sundaresan2020dense,
  author    = {Priya Sundaresan and Jennifer Grannen and Brijen Thananjeyan and Ashwin Balakrishna and Michael Laskey and Kevin Stone and Joseph E. Gonzalez and Ken Goldberg},
  title     = {Learning Rope Manipulation Policies Using Dense Object Descriptors Trained on Synthetic Depth Data},
  booktitle = {2020 IEEE International Conference on Robotics and Automation (ICRA)},
  year      = {2020},
  pages     = {9411--9418},
  doi       = {10.1109/ICRA40945.2020.9197121},
  url       = {https://doi.org/10.1109/ICRA40945.2020.9197121}
}

@inproceedings{wilson2023cable,
  author    = {Achu Wilson and Helen Jiang and Wenzhao Lian and Wenzhen Yuan},
  title     = {Cable Routing and Assembly Using Tactile-Driven Motion Primitives},
  booktitle = {2023 IEEE International Conference on Robotics and Automation (ICRA)},
  year      = {2023},
  pages     = {10408--10414},
  doi       = {10.1109/ICRA48891.2023.10161069},
  url       = {https://doi.org/10.1109/ICRA48891.2023.10161069}
}

@inproceedings{viswanath2021disentangling,
  author    = {Vainavi Viswanath and Jennifer Grannen and Priya Sundaresan and Brijen Thananjeyan and Ashwin Balakrishna and Ellen Novoseller and Jeffrey Ichnowski and Michael Laskey and Joseph E. Gonzalez and Ken Goldberg},
  title     = {Disentangling Dense Multi-Cable Knots},
  booktitle = {2021 IEEE/RSJ International Conference on Intelligent Robots and Systems (IROS)},
  year      = {2021},
  pages     = {3731--3738},
  doi       = {10.1109/IROS51168.2021.9636397},
  url       = {https://doi.org/10.1109/IROS51168.2021.9636397}
}

@inproceedings{viswanath2023handloom,
  author    = {Vainavi Viswanath and Kaushik Shivakumar and Mallika Parulekar and Jainil Ajmera and Justin Kerr and Jeffrey Ichnowski and Richard Cheng and Thomas Kollar and Ken Goldberg},
  title     = {{HANDLOOM}: Learned Tracing of One-Dimensional Objects for Inspection and Manipulation},
  booktitle = {Proceedings of the 7th Conference on Robot Learning},
  year      = {2023},
  volume    = {229},
  series    = {Proceedings of Machine Learning Research},
  pages     = {341--357},
  publisher = {PMLR},
  url       = {https://proceedings.mlr.press/v229/viswanath23a.html}
}

@inproceedings{seita2021gctn,
  author    = {Daniel Seita and Pete Florence and Jonathan Tompson and Erwin Coumans and Vikas Sindhwani and Ken Goldberg and Andy Zeng},
  title     = {Learning to Rearrange Deformable Cables, Fabrics, and Bags with Goal-Conditioned Transporter Networks},
  booktitle = {2021 IEEE International Conference on Robotics and Automation (ICRA)},
  year      = {2021},
  pages     = {4568--4575},
  doi       = {10.1109/ICRA48506.2021.9561391},
  url       = {https://doi.org/10.1109/ICRA48506.2021.9561391}
}

@inproceedings{kim2024openvla,
  author    = {Moo Jin Kim and Karl Pertsch and Siddharth Karamcheti and Ted Xiao and Ashwin Balakrishna and Suraj Nair and Rafael Rafailov and Ethan P. Foster and Pannag R. Sanketi and Quan Vuong and Thomas Kollar and Benjamin Burchfiel and Russ Tedrake and Dorsa Sadigh and Sergey Levine and Percy Liang and Chelsea Finn},
  title     = {{OpenVLA}: An Open-Source Vision-Language-Action Model},
  booktitle = {Proceedings of the 8th Conference on Robot Learning},
  year      = {2025},
  editor    = {Pulkit Agrawal and Oliver Kroemer and Wolfram Burgard},
  volume    = {270},
  series    = {Proceedings of Machine Learning Research},
  pages     = {2679--2713},
  publisher = {PMLR},
  note      = {Conference held in 2024},
  url       = {https://proceedings.mlr.press/v270/kim25c.html}
}

@inproceedings{black2025pi0,
  title = {{$\pi_0$}: A Vision-Language-Action Flow Model
           for General Robot Control},
  author = {Kevin Black and Noah Brown and Danny Driess and
            Adnan Esmail and Michael Robert Equi and Chelsea Finn and
            Niccolo Fusai and Lachy Groom and Karol Hausman and
            Brian Ichter and Szymon Jakubczak and Tim Jones and
            Liyiming Ke and Sergey Levine and Adrian Li-Bell and
            Mohith Mothukuri and Suraj Nair and Karl Pertsch and
            Lucy Xiaoyang Shi and Laura Smith and James Tanner and
            Quan Vuong and Anna Walling and Haohuan Wang and
            Ury Zhilinsky},
  booktitle = {Proceedings of Robotics: Science and Systems},
  year = {2025},
  doi = {10.15607/RSS.2025.XXI.010},
  url = {https://www.roboticsproceedings.org/rss21/p010.html}
}

@misc{black2025pi05,
      title={$\pi_{0.5}$: a Vision-Language-Action Model with Open-World Generalization}, 
      author={{Physical Intelligence} and Kevin Black and Noah Brown and James Darpinian and Karan Dhabalia and Danny Driess and Adnan Esmail and Michael Equi and Chelsea Finn and Niccolo Fusai and Manuel Y. Galliker and Dibya Ghosh and Lachy Groom and Karol Hausman and Brian Ichter and Szymon Jakubczak and Tim Jones and Liyiming Ke and Devin LeBlanc and Sergey Levine and Adrian Li-Bell and Mohith Mothukuri and Suraj Nair and Karl Pertsch and Allen Z. Ren and Lucy Xiaoyang Shi and Laura Smith and Jost Tobias Springenberg and Kyle Stachowicz and James Tanner and Quan Vuong and Homer Walke and Anna Walling and Haohuan Wang and Lili Yu and Ury Zhilinsky},
      year={2025},
      eprint={2504.16054},
      archivePrefix={arXiv},
      primaryClass={cs.LG},
      url={https://arxiv.org/abs/2504.16054}, 
}

@inproceedings{radford2021clip,
  author    = {Alec Radford and Jong Wook Kim and Chris Hallacy and Aditya Ramesh and Gabriel Goh and Sandhini Agarwal and Girish Sastry and Amanda Askell and Pamela Mishkin and Jack Clark and Gretchen Krueger and Ilya Sutskever},
  title     = {Learning Transferable Visual Models From Natural Language Supervision},
  booktitle = {Proceedings of the 38th International Conference on Machine Learning},
  year      = {2021},
  editor    = {Marina Meila and Tong Zhang},
  volume    = {139},
  series    = {Proceedings of Machine Learning Research},
  pages     = {8748--8763},
  publisher = {PMLR},
  url       = {https://proceedings.mlr.press/v139/radford21a.html}
}

@inproceedings{carion2025sam3,
  author    = {Nicolas Carion and Laura Gustafson and Yuan-Ting Hu and Shoubhik Debnath and Ronghang Hu and Didac Suris Coll-Vinent and Chaitanya Ryali and Kalyan Vasudev Alwala and Haitham Khedr and Andrew Huang and Jie Lei and Tengyu Ma and Baishan Guo and Arpit Kalla and Markus Marks and Joseph Greer and Meng Wang and Peize Sun and Roman R{\"a}dle and Triantafyllos Afouras and Effrosyni Mavroudi and Katherine Xu and Tsung-Han Wu and Yu Zhou and Liliane Momeni and Rishi Hazra and Shuangrui Ding and Sagar Vaze and Francois Porcher and Feng Li and Siyuan Li and Aishwarya Kamath and Ho Kei Cheng and Piotr Doll{\'a}r and Nikhila Ravi and Kate Saenko and Pengchuan Zhang and Christoph Feichtenhofer},
  title     = {{SAM 3}: Segment Anything with Concepts},
  booktitle = {International Conference on Learning Representations},
  year      = {2026},
  eprint    = {2511.16719},
  archivePrefix = {arXiv},
  primaryClass  = {cs.CV},
  url       = {https://proceedings.iclr.cc/paper_files/paper/2026/hash/e0982cbc81401df3430ee1ff780dc7a2-Abstract-Conference.html}
}

@inproceedings{li2026hierarchical,
  author        = {Mingen Li and Houjian Yu and Yixuan Huang and Youngjin Hong and Hantao Ye and Changhyun Choi},
  title         = {Hierarchical {DLO} Routing with Reinforcement Learning and In-Context Vision-Language Models},
  booktitle     = {2026 IEEE International Conference on Robotics and Automation (ICRA)},
  year          = {2026},
  note          = {Finalist for the ICRA 2026 Best Paper Award on Robot Learning},
  eprint        = {2510.19268},
  archivePrefix = {arXiv},
  primaryClass  = {cs.RO},
  url           = {https://icra2026-dloroute.github.io/DLORoute/}
}

@inproceedings{cao2026dlolab,
  author        = {Junyi Cao and Yian Wang and Ziyan Xiong and Chunru Lin and Zhehuan Chen and Chuang Gan},
  title         = {{DLO-Lab}: Benchmarking Deformable Linear Object Manipulations with Differentiable Physics},
  booktitle     = {Proceedings of the 43rd International Conference on Machine Learning (ICML)},
  year          = {2026},
  eprint        = {2606.04206},
  archivePrefix = {arXiv},
  primaryClass  = {cs.RO},
  url           = {https://icml.cc/virtual/2026/poster/63391}
}

@misc{wirecraft2026,
  author        = {Chongyu Zhu and Ramy ElMallah and Hyegang Kim and Zachary Tang and Jiachen Rao and Artem Arutyunov and Seungyeon Ha and Chi-Guhn Lee},
  title         = {{WireCraft}: A Simulation Benchmark for Industrial {DLO} Manipulation},
  year          = {2026},
  eprint        = {2606.18097},
  archivePrefix = {arXiv},
  primaryClass  = {cs.RO},
  doi           = {10.48550/arXiv.2606.18097},
  url           = {https://arxiv.org/abs/2606.18097}
}

@article{liu2026sam3dvla,
  title = {{SAM3D}-Guided Object-Centric Representation Alignment
           for Vision-Language-Action Models},
  author = {Zonghe Liu and Shanyuan Jie and Xiaoquan Sun and
            Chen Cao and Zetian Xu and Zongsheng Liu and Jiayu Chen},
  journal = {arXiv preprint arXiv:2607.25912},
  year = {2026},
  doi = {10.48550/arXiv.2607.25912},
  url = {https://arxiv.org/abs/2607.25912}
}

@article{affordvla2026,
  title = {{AffordVLA}: Injecting Affordance Representations into
           Vision-Language-Action Models via Implicit Feature Alignment},
  author = {Weijie Kong and Zhian Su and Wei Yu and Huixu Dong},
  journal = {arXiv preprint arXiv:2605.17517},
  year = {2026},
  doi = {10.48550/arXiv.2605.17517},
  url = {https://arxiv.org/abs/2605.17517}
}

@article{lou2026mwm,
  title = {Mask World Model: Predicting What Matters for
           Robust Robot Policy Learning},
  author = {Yunfan Lou and Xiaowei Chi and Xiaojie Zhang and
            Zezhong Qian and Chengxuan Li and Rongyu Zhang and
            Yaoxu Lyu and Guoyu Song and Chuyao Fu and Haoxuan Xu and
            Pengwei Wang and Shanghang Zhang},
  journal = {arXiv preprint arXiv:2604.19683},
  year = {2026},
  doi = {10.48550/arXiv.2604.19683},
  url = {https://arxiv.org/abs/2604.19683}
}

@article{li2026efficientwam,
  title = {{Efficient-WAM}: A 1{B}-Parameter World-Action Model
           with Low-Cost Future Imagination},
  author = {Jiajun Li and Tiecheng Guo and Yifan Ye and
            Rongyu Zhang and Xiaowei Chi and Qianpu Sun and
            Ying Li and Yunfan Lou and Yan Huang and Zhihe Lu and
            Meng Guo and Shanghang Zhang},
  journal = {arXiv preprint arXiv:2606.10040},
  year = {2026},
  doi = {10.48550/arXiv.2606.10040},
  url = {https://arxiv.org/abs/2606.10040}
}

@inproceedings{zheng2025flare,
  title = {{FLARE}: Robot Learning with Implicit World Modeling},
  author = {Ruijie Zheng and Jing Wang and Scott Reed and
            Johan Bjorck and Yu Fang and Fengyuan Hu and
            Joel Jang and Kaushil Kundalia and Zongyu Lin and
            Lo{\"i}c Magne and Avnish Narayan and You Liang Tan and
            Guanzhi Wang and Qi Wang and Jiannan Xiang and
            Yinzhen Xu and Seonghyeon Ye and Jan Kautz and
            Furong Huang and Yuke Zhu and Linxi Fan},
  booktitle = {Proceedings of the 9th Conference on Robot Learning},
  series = {Proceedings of Machine Learning Research},
  volume = {305},
  pages = {3952--3971},
  publisher = {PMLR},
  year = {2025},
  url = {https://proceedings.mlr.press/v305/zheng25a.html}
}

@misc{rdt,
      title={R{DT}-1{B}: a Diffusion Foundation Model for Bimanual Manipulation}, 
      author={Songming Liu and Lingxuan Wu and Bangguo Li and Hengkai Tan and Huayu Chen and Zhengyi Wang and Ke Xu and Hang Su and Jun Zhu},
      year={2025},
      eprint={2410.07864},
      archivePrefix={arXiv},
      primaryClass={cs.RO},
      url={https://arxiv.org/abs/2410.07864}, 
}

@misc{fast,
      title={{FAST}: Efficient Action Tokenization for Vision-Language-Action Models}, 
      author={Karl Pertsch and Kyle Stachowicz and Brian Ichter and Danny Driess and Suraj Nair and Quan Vuong and Oier Mees and Chelsea Finn and Sergey Levine},
      year={2025},
      eprint={2501.09747},
      archivePrefix={arXiv},
      primaryClass={cs.RO},
      url={https://arxiv.org/abs/2501.09747}, 
}

@misc{LingBotVLA2,
      title={From Foundation to Application: Improving {VLA} Models in Practice}, 
      author={Wei Wu and Fangjing Wang and Fan Lu and He Sun and Shi Liu and Yunnan Wang and Yibin Yan and Yong Wang and Shuailei Ma and Xinyang Wang and Yibin Liu and Shuai Yang and Tianxiang Zhou and Kejia Zhang and Lei Zhou and Cheng Su and Nan Xue and Bin Tan and Han Zhang and Youchao Zhang and Fei Liao and Xing Zhu and Yujun Shen and Kecheng Zheng},
      year={2026},
      eprint={2607.06403},
      archivePrefix={arXiv},
      primaryClass={cs.RO},
      url={https://arxiv.org/abs/2607.06403}, 
}

@misc{n1.7,
      title={{GR00T} {N}1: An Open Foundation Model for Generalist Humanoid Robots}, 
      author={NVIDIA and : and Johan Bjorck and Fernando Castañeda and Nikita Cherniadev and Xingye Da and Runyu Ding and Linxi "Jim" Fan and Yu Fang and Dieter Fox and Fengyuan Hu and Spencer Huang and Joel Jang and Zhenyu Jiang and Jan Kautz and Kaushil Kundalia and Lawrence Lao and Zhiqi Li and Zongyu Lin and Kevin Lin and Guilin Liu and Edith Llontop and Loic Magne and Ajay Mandlekar and Avnish Narayan and Soroush Nasiriany and Scott Reed and You Liang Tan and Guanzhi Wang and Zu Wang and Jing Wang and Qi Wang and Jiannan Xiang and Yuqi Xie and Yinzhen Xu and Zhenjia Xu and Seonghyeon Ye and Zhiding Yu and Ao Zhang and Hao Zhang and Yizhou Zhao and Ruijie Zheng and Yuke Zhu},
      year={2025},
      eprint={2503.14734},
      archivePrefix={arXiv},
      primaryClass={cs.RO},
      url={https://arxiv.org/abs/2503.14734}, 
}

@misc{dreamzero,
      title={World Action Models are Zero-shot Policies}, 
      author={Seonghyeon Ye and Yunhao Ge and Kaiyuan Zheng and Shenyuan Gao and Sihyun Yu and George Kurian and Suneel Indupuru and You Liang Tan and Chuning Zhu and Jiannan Xiang and Ayaan Malik and Kyungmin Lee and William Liang and Nadun Ranawaka and Jiasheng Gu and Yinzhen Xu and Guanzhi Wang and Fengyuan Hu and Avnish Narayan and Johan Bjorck and Jing Wang and Gwanghyun Kim and Dantong Niu and Ruijie Zheng and Yuqi Xie and Jimmy Wu and Qi Wang and Ryan Julian and Danfei Xu and Yilun Du and Yevgen Chebotar and Scott Reed and Jan Kautz and Yuke Zhu and Linxi "Jim" Fan and Joel Jang},
      year={2026},
      eprint={2602.15922},
      archivePrefix={arXiv},
      primaryClass={cs.RO},
      url={https://arxiv.org/abs/2602.15922}, 
}

@misc{openvla-oft,
      title={Fine-Tuning Vision-Language-Action Models: Optimizing Speed and Success}, 
      author={Moo Jin Kim and Chelsea Finn and Percy Liang},
      year={2025},
      eprint={2502.19645},
      archivePrefix={arXiv},
      primaryClass={cs.RO},
      url={https://arxiv.org/abs/2502.19645}, 
}

@inproceedings{tarvainen2017mean,
  title     = {Mean Teachers Are Better Role Models:
               Weight-Averaged Consistency Targets Improve
               Semi-Supervised Deep Learning Results},
  author    = {Tarvainen, Antti and Valpola, Harri},
  booktitle = {Advances in Neural Information Processing Systems},
  volume    = {30},
  year      = {2017},
  url       = {https://proceedings.neurips.cc/paper/2017/hash/68053af2923e00204c3ca7c6a3150cf7-Abstract.html}
}

@inproceedings{zhu2023learning,
  title     = {Learning Generalizable Manipulation Policies
               with Object-Centric 3{D} Representations},
  author    = {Zhu, Yifeng and Jiang, Zhenyu and
               Stone, Peter and Zhu, Yuke},
  booktitle = {Proceedings of the 7th Conference on Robot Learning},
  series    = {Proceedings of Machine Learning Research},
  volume    = {229},
  pages     = {3418--3433},
  year      = {2023},
  publisher = {PMLR},
  url       = {https://proceedings.mlr.press/v229/zhu23b.html}
}

@article{yao2026manipulation,
  title={Manipulation Degree: A Novel Metric to Evaluate the Maneuverability of Magnetic Guidewires},
  author={Yao, Bo-Xian and Zhou, Xiao-Hu and Gui, Mei-Jiang and Xie, Xiao-Liang and Liu, Shi-Qi and Wang, Shuang-Yi and Li, Guo-Tao and Tang, Yong-Qiang and Li, Ying-Tian and Hou, Zeng-Guang},
  journal={IEEE Transactions on Medical Robotics and Bionics},
  year={2026},
  publisher={IEEE}
}

@article{zhang2025large,
  title={How large AI model empowers time-series forecasting for the operation and maintenance of industrial automation system?},
  author={Zhang, Le and Cheng, Wei and Zhang, Shuo and Xing, Ji and Nie, Zelin and Chen, Xuefeng and Lan, Dapeng and Liu, Yu and Yang, Yun and Pang, Zhibo},
  journal={IEEE Transactions on Industrial Informatics},
  year={2025},
  publisher={IEEE}
}

\end{document}